\RequirePackage[log]{snapshot}
\documentclass[10pt,twocolumn,letterpaper]{article}

\usepackage{cvpr}
\usepackage{times}
\usepackage{epsfig}
\usepackage{graphicx}
\usepackage{amsmath}
\usepackage{amssymb}
\usepackage{fancyhdr}

\usepackage[pagebackref=true,breaklinks=true,letterpaper=true,colorlinks,bookmarks=false]{hyperref}

\usepackage{graphics} 
\usepackage{url}      
\usepackage{subfigure}
\usepackage{float}
\usepackage{array}

\newcommand{\parahead}[1]{\par\textbf{#1}:\ }
\newcommand{\tabhead}[1]{\par\textbf{#1}}

\newcommand{\E}{E}

\newcommand{\MU}{\boldsymbol{\mu}}
\newcommand{\C}{\mathcal{C}}
\newcommand{\G}{\mathcal{G}}

\cvprfinalcopy 

\def\cvprPaperID{0123} 

\ifcvprfinal\fi

\usepackage{fancyhdr}
\begin{document}


\title{Fast and Robust Hand Tracking Using Detection-Guided Optimization}

\author{Srinath Sridhar\textsuperscript{1}, Franziska Mueller\textsuperscript{1,2}, 
Antti Oulasvirta\textsuperscript{3}, Christian Theobalt\textsuperscript{1}\\
\textsuperscript{1}Max Planck Institute for Informatics, \textsuperscript{2}Saarland University, \textsuperscript{3}Aalto University\\
{\tt\small \{ssridhar,frmueller,theobalt\}@mpi-inf.mpg.de, antti.oulasvirta@aalto.fi}
}


\maketitle
\thispagestyle{empty} 
\thispagestyle{fancy}

\begin{abstract}
  %
  Markerless tracking of hands and fingers is a promising enabler for human-computer interaction. 
  However, adoption has been limited because of tracking inaccuracies, incomplete coverage of motions, low framerate, complex camera setups,
  and high computational requirements.
  In this paper, we present a fast method for accurately tracking rapid and complex articulations of the hand using a single depth camera.
  Our algorithm uses a novel \textbf{detection-guided optimization} strategy that increases the robustness and speed of pose estimation.
  In the detection step, a randomized decision forest classifies pixels into parts of the hand.
  In the optimization step, a novel objective function combines the detected part labels and
  a Gaussian mixture representation of the depth to estimate a pose that best fits the depth.
  Our approach needs comparably less computational resources which makes it extremely fast (50~fps without GPU support).
  The approach also supports varying static, or moving, \mbox{camera-to-scene} arrangements.
  We show the benefits of our method by evaluating on public datasets and comparing against previous work.
\end{abstract}

\section{Introduction}
There is increasing interest in using markerless hand tracking in human-computer interaction, for instance when
interacting with 3D applications, augmented reality, smart watches, and for gestural input~\cite{kim_digits:_2012,lee_spacetop:_2013,wang_6d_2011}.
However, flexible, realtime markerless tracking of hands presents several unique challenges.
First, natural hand movement involves simultaneous control of several ($\geq$ 25) degrees-of-freedom (DOFs), fast motions with rapid changes in direction, and self-occlusions.
Tracking fast and complex \emph{finger articulations} combined with global motion of the hand at \emph{high framerates} is critical but remains a challenging problem.
Second, many methods use dense camera setups~\cite{oikonomidis_full_2011,sridhar_interactive_2013} or GPU acceleration~\cite{oikonomidis_efficient_2011}, \ie have high \emph{setup costs} which limits deployment.
Finally, applications of hand tracking demand tracking across many camera-to-scene configurations including desktop, egocentric and wearable settings.

This paper presents a novel method for hand tracking with a single depth camera that aims to address these challenges.
Our method is extremely fast (nearly equalling the capture rate of the camera), reliable, and supports varying close-range camera-to-hand arrangements
including desktop, and moving egocentric (camera mounted to the head).

\begin{figure*}[ht!]
  \centering
  \includegraphics[width=0.9\textwidth]{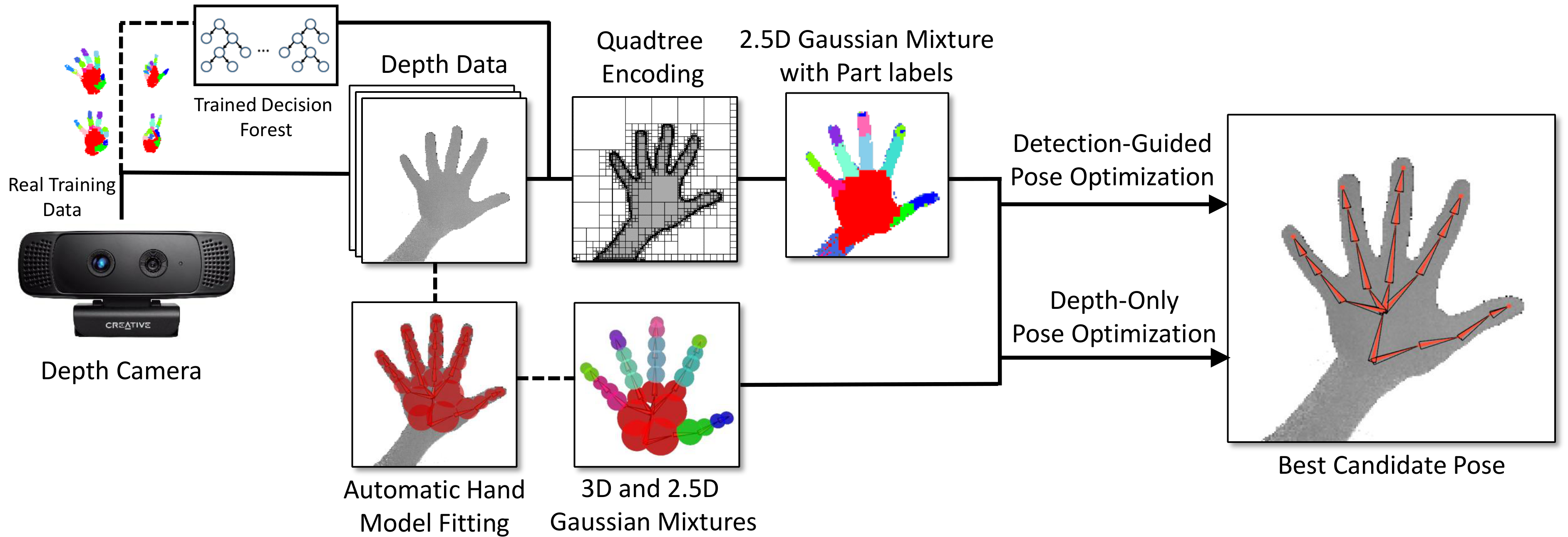}
  \caption{Overview of our detection-guided tracking method. We develop a novel representation for depth data and hand model as a mixture of 2.5D Gaussians. This representation allows us to combine the benefits of model-based generative tracking and discriminative part detection. Pixels classified using a trained decision forest are directly incorporated as evidence in detection-guided pose optimization. Dashed lines indicate offline computation. Best viewed in color.}
  \label{fig:pipeline}
\end{figure*}

The main novelty in our work is a new \textbf{detection-guided optimization} strategy that 
combines the benefits of two common strands in hand tracking research---model-based generative tracking and discriminative hand pose detection---into a unified framework that yields high efficiency and robust performance and minimizes their mutual failures (see Figure~\ref{fig:pipeline}). 
The first contribution in this strategy is a novel, efficient representation of both the input depth and the hand model shape as a mixture of Gaussian functions.
While previous work used primitive shapes like cylinders~\cite{oikonomidis_efficient_2011,oikonomidis_full_2011} or spheres~\cite{qian_realtime_2014} to represent
the hand model, we use Gaussian mixtures for both the depth data and the model.
This compact, mathematically smooth representation allows us to formulate pose estimation as a 2.5D generative optimization problem in depth.
We define a new \textbf{depth-only} energy, that optimizes for the similarity of the input depth with the hand model.
It uses additional prior and data terms to avoid finger collisions and preserve the smoothness of reconstructed motions.
Importantly, since the energy is smooth, we can obtain analytic gradients and perform rapid optimization. 
While pose tracking on this energy alone could run in excess of 120~fps using gradient-based local optimization, this often results in a wrong local pose optimum.
 
The second contribution in our strategy is thus to incorporate evidence from trained randomized decision forests that label depth pixels into predefined parts of the hand.
Unlike previous purely detection-based approaches~\cite{Fanello_2014,shotton_real-time_2011}, we use the part labels as additional constraints in an augmented version of the aforementioned depth-only energy, henceforth termed \textbf{detection-guided} energy.
The part labels include discriminative detection evidence into generative pose estimation.
This enables the tracker to better recover from erroneous local pose optima and prevents temporal jitter common to detection-only approaches.
The precondition for recovery is reliability of the part labels.
However, even with large training sets it is hard to obtain perfect part classification (per-pixel accuracy is usually around 60\%).
Thus, pose estimation based on this additional discriminative evidence is also not sufficient.

Our third contribution therefore, is a new \textbf{late fusion approach} that combines particle-based multi-hypothesis optimization with an efficient local gradient-based optimizer.
Previous work has used particle-based optimizers, but they tend to be computationally expensive~\cite{oikonomidis_efficient_2011,oikonomidis_full_2011}.
Our approach is fast because we combine the speed of local gradient-based optimization with the robustness of particle-based approaches.
At each time step of depth video, a set of initial pose hypotheses (particles) is generated, from which a subsequent local optimization is started.
Some of these local optimizers use the depth-only pose energy, some others use the detection-guided energy.
In a final late fusion step the best pose is chosen based on the pose fitting energy.

Our approach results in a temporally stable and efficient tracker that estimates full articulated joint angles of
even rapid and complex hand motions at previously unseen frame rates in excess of 50~fps, even with a CPU implementation.
Our tracker is resilient to erroneous local convergence by resorting to the detection-guided solution when labels can be trusted, and it
is not misled by erroneous detections as it can then switch to the depth-only tracking result.

We show these improvements with (1) qualitative experiments, (2) extensive evaluation on public datasets, and (3) comparisons
with other state-of-the-art methods.

\section{Related Work}
In this brief review we focus on previous approaches to markerless hand tracking from depth images.
First, we briefly discuss marker-based and multi-camera techniques.
Gloves fitted with retro-reflective markers or color patches were used to estimate the
kinematic skeleton using inverse kinematics~\cite{sturman_survey_1994,wang_real-time_2009,zimmerman_hand_1986}.
Research on \emph{markerless} tracking was made popular in the early 2000s (\eg~\cite{athitsos_estimating_2003,wu_capturing_2001}). 
Some recent solutions assume a multi-camera setup with offline
processing~\cite{ballan_motion_2012,oikonomidis_full_2011,wang_video-based_2013}, 
while others track at interactive rates~\cite{sridhar_interactive_2013,wang_6d_2011} of up to 30~fps~\cite{sridhar2014real}.
However, calibrated multi-camera setups make these methods difficult to adopt for practical applications.
The recent introduction of consumer depth sensors has resulted in a number of methods that require only a single depth camera.
Some commercial solutions exist, such as the Leap Motion.
Although Leap Motion is fast, the approach uses strong priors and fails with complex self-occlusions and non-standard motions (we show an example in Section~\ref{sec:results}).

The main approaches to real-time hand tracking can be divided into two classes: (1) generative and (2) discriminative methods.\footnote{There are algorithmic
parallels to full-body tracking~\cite{baak_data-driven_2011,hutchison_real-time_2012,kurmankhojayev_monocular_2013,shotton_real-time_2011}.}
First, \cite{hamer_tracking_2009} proposed a method to track a hand manipulating an object that takes 6.2~s/frame.
\cite{oikonomidis_efficient_2011} proposed a model-based method that made use of particle-swarm optimization.
This method requires GPU acceleration to achieve 15~fps and uses skin color segmentation which is sensitive to lighting.
They showed an extension to interacting hands, although only offline~\cite{Oikonomidis:CVPR2012,oikonomidis_evolutionary_2014}.
\cite{melax_dynamics_2013} proposed a tracking method directly in depth by efficient parallel physics simulations.
While this method is fast, finger articulations are often incorrectly tracked, as we demonstrate later.
Recent real-time surface tracking methods from depth~\cite{Zollhofer:2014} were applied to hands, but are limited to simple motions with no occlusions.

Second, decision forests were used with great success for full body tracking~\cite{girshick_efficient_2011,shotton_real-time_2011} and later adopted to hand tracking with varying success.
\cite{keskin_real_2011} proposed a method for recognizing finger spelling in depth data using classification forests.
\cite{Fanello_2014,tang_latent_2014,tang_real-time_2013,xu_efficient_2013} also proposed methods based on variants of random forests.
Tompson \etal~\cite{Tompson_TOG_2014} track hand motion from depth at $\leq$ 25~fps using feature detections from a
convolutional network and further pose refinement through inverse kinematics.
However, a common problem with these approaches is jitter due to missing temporal information at each time step.
We provide a direct comparison with one recent method~\cite{tang_latent_2014} to demonstrate this.
Moreover, most methods estimate joint positions with temporally varying bone lengths, limiting applicability.

\cite{sridhar_interactive_2013} proposed combining discriminative and generative hand pose estimation.
Their approach detected only fingertips, which could easily be occluded or misdetected.
Offline tracking in RGBD using a combination of discriminative and generative pose estimation was shown in~\cite{Gall_GCPR_2014}.
\cite{qian_realtime_2014} proposed a method based on optimization in combination with discriminative fingertip detection, achieving 25~fps.
However, tracking would be hard with this method when one or more of the fingertips are occluded.

In this paper we present a method that combines decision forests and pose estimation in a unified optimization framework.
To our knowledge, ours is the first method to track rapid articulations at 50~fps using a single depth camera and yet achieve state-of-the-art accuracy.

\section{Input and Model Representation}\label{sec:input}
In the past, representations such as spheres or cylinders have been used to represent the hand model~\cite{oikonomidis_efficient_2011,qian_realtime_2014}.
Similarly, downsampled images~\cite{wang_6d_2011,wang_real-time_2009} or silhouettes~\cite{ballan_motion_2012} have been used as representations of input data.
However, such representations make pose optimization energies discontinuous and difficult to optimize.
Our novel representation of depth and 3D model data uses a mixture of \emph{weighted} Gaussian functions to
represent both depth data and the hand shape.
We were inspired by~\cite{stoll_fast_2011} who use multiple 2D RGB images and \cite{kurmankhojayev_monocular_2013} who use depth data.
Both methods rely on a uniformly weighted Gaussian mixture, and a 2D or 3D error metric for pose estimation.
However, we make important modifications that allows representing 3D depth data using a 2.5D formulation since data from depth sensors contains
information only about the \emph{camera-facing} parts of the scene.
Thus, we enable pose estimation based on alignment to a single depth image using a 2.5D error metric.

An instance of the input depth or the hand model can be represented as a mixture of Gaussian functions
\begin{align}
  \mathcal{\C}(\mathbf{x}) = \sum_{i=1}^n w_i \, \G_i(\mathbf{x}; \sigma, \MU),
  \label{eqn:gmm}
\end{align}
where $\G_i(.)$ denotes a unnormalized Gaussian function with isotropic variance, $\sigma^2$, in all dimensions of $\mathbf{x} \in \mathcal{R}^n$, and mean $\MU$.
%
The Gaussian mixture representation has many advantages.
First, it enables a mathematically smooth pose estimation energy which is analytically differentiable. 
Second, only a few Gaussians are needed for representing the input depth and the hand model, an implicit data reduction which makes optimization extremely fast.
Finally, it provides a natural way to compute collisions using an analytically differentiable energy.
We show later that collisions are important for pose estimation (Section~\ref{sec:tracking}).
To aid visualization we henceforth represent each Gaussian in the mixture as a sphere ($\mathbf{x} \in R^3$) or circle ($\mathbf{x} \in R^2$) with a radius of $1 \, \sigma$.  
However, Gaussians have infinite support ($\C(\mathbf{x}) > 0 $ everywhere) and can produce long range attractive or repulsive \emph{force} during pose optimization.

\subsection{Depth Data Representation}\label{sec:depth_rep}
The depth camera outputs depth maps, \ie each pixel has an associated depth value.
Depth maps contain only the camera-facing parts of the scene and information about occluded parts is unavailable.
We therefore only represent the camera-facing pixels using Gaussian mixtures, which are computed in real-time. 

First, we decompose the depth image into regions of homogeneous depth using a quadtree.
The quadtree recursively decomposes depth image regions further, until the depth difference between
the furthest and nearest point in a region is below a threshold $\epsilon_c$ ($\epsilon_c = 20$~mm in all experiments).
%
To each quad in the tree, we fit a Gaussian function with $\MU$ set to the center of the quad, and $\sigma = {a}/\sqrt{2}$, where $a$ is the side length of the quad.
We also set each Gaussian function to have unit weight $w_i$ since we consider all input data to be equally important.
This leads us to an analytic representation of the \emph{camera-facing surface} of the input depth, $\C_I(\mathbf{x}) = \sum_{q=1}^n \G_q(\mathbf{x})$,
where $\mathbf{x} \in \mathcal{R}^2$ and $n$ is the number of leaves in the quadtree.
Additionally, each quad has an associated depth value, $d_q$, which is the mean of all depth pixels within the quad.
Figure~\ref{fig:pipeline} illustrates the process of converting input depth to a Gaussian mixture.

\subsection{Hand Model}
We model the volumetric extent of the hand analytically using a mixture of 3D Gaussian functions, \mbox{$\C_h(\mathbf{x}) = \sum_{h=1}^m w_h \, \G_h(\mathbf{x})$}
where $\mathbf{x} \in \mathcal{R}^3$ and $m$ is the number of Gaussians.
We assume that the best fitting model has Gaussians whose isosurface at $1 \, \sigma$ coincides with the surface of the hand.
In Section~\ref{sec:handmodeling} we present a fully automatic procedure to fit such a hand model to a user.
Additionally, $\C_h$, is attached to a \emph{parametric}, kinematic skeleton similar to that of~\cite{simo_serra_kinematic_2011}, \ie each 3D Gaussian is attached to 
a bone which determines its mean position in 3D. 
We use $|\Theta| = 26$ skeletal pose parameters in twist representation, including 3 translational DOFs, 3 global rotations, and 20 joint angles.

\parahead{Model Surface Representation}
$\C_I$ is a representation of the \emph{camera-facing surface} while $\C_h$ represents the full volumetric extent of the hand.
In order to create an equivalent representation of the hand model that approximates the camera-facing parts, which we later use in pose optimization (Section~\ref{sec:tracking}). 
For each model Gaussian in $\C_h$, we create a new projected Gaussian such that the projected hand model has the form $\C_p = \sum_{p=1}^m w_p \, \G_p(\mathbf{x})$
where $\mathbf{x} \in R^2$ and $w_p = w_h \, \forall \, h$.
$\C_p$ is a representation of the hand model as seen from the perspective of the depth camera and is defined over the depth image domain.
The parameters of each Gaussian $\G_p$ are set to be $(\MU_p, \sigma_p)$,
where $\MU_p = \mathbf{K} \, [ \,\mathbf{I} \, |\, \mathbf{0} \,] \, \MU_h$.
Like~\cite{stoll_fast_2011} we approximate the perspective projection with a scaled orthographic projection, yielding 2D Gaussians 
with $\sigma_p = \sigma_h \, f / \left[\MU_p\right]_z$.
Here $f$ is the focal length of the camera, and $\left[\MU_p\right]_z$ denotes the $z$-coordinate of the Gaussian mean.


\section{Hand Pose Optimization}\label{sec:tracking}
In this section we describe our new formulation of pose estimation as an optimization problem using the Gaussian mixture representation of 2.5D depth data (See Figure~\ref{fig:pipeline}).
Our algorithm uses two variants of a model-to-image similarity energy, one that is 
only based on depth data (Section~\ref{sec:depth_based_pose}), and another that is guided by decision forests-based part detection (Section~\ref{sec:detection_guided_pose}). 
Pose estimates obtained with each energy are used by a late fusion approach to find the final pose estimate (Section~\ref{sec:fusion}).
Input to pose optimization at each time step of depth video is the 2.5D mixture of Gaussians representation of a depth image  $\C_I$.
The latter is computed after median filtering the depth (to remove flying pixels in time-of-flight data), and for a constrained working volume in 
depth between 150~mm and 600~mm from the camera. The 3D Gaussian mixture of the hand model is denoted by $\C_h$
and its projected version is denoted by $\C_p$.

\subsection{Depth-Only Pose Optimization}\label{sec:depth_based_pose}
Our goal is to optimize for the skeleton pose parameters $\Theta$ that best explain the input data and are anatomically plausible.
We frame an energy that satisfies our goal while being mathematically smooth and differentiable. These properties make the energy ideal for fast optimization.

\subsection{Objective Function}\label{sec:objective}
Our new energy has the following general form:
\begin{align}
  \mathcal{E}(\Theta) &= E_{sim} - w_c \, E_{col} \nonumber\\&- w_l \, E_{lim} - w_s \, E_{smo},
  \label{eqn:objective}
\end{align}
where $E_{sim}$ is a measure of 2.5D similarity between $\C_I$ and $\C_p$,
$E_{col}$ is a penalty for collisions between Gaussians in $\C_h$,
$E_{lim}$ enforces a soft constraint on the skeleton joint limits,
$E_{smo}$ enforces smoothness in the tracked motion.
In all our experiments, we used fixed weighting factors chosen by searching for the best accuracy over the dataset: $w_c = 1.0$, $w_l = 0.2$, and $w_s = 1.0$.
Before describing each of the terms in detail we first introduce a measure of similarity between two Gaussian mixtures which is the basis for many of the terms in the objective.

\parahead{Gaussian Similarity Measure}
We define a similarity measure between any two pairs of Gaussian mixtures $\C_a$ and $\C_b$ as,
\begin{align}
  E(\C_a,\C_b)  \label{eqn:gausssim}
  & = \sum_{p \in \C_a} \sum_{q \in \C_b} D_{pq},\\
  \text{where}, D_{pq} & = \, w_p \, w_q \, \int_\Omega \G_p(\mathbf{x}) \, \G_q(\mathbf{x}) \, \mathrm{d}\mathbf{x},\label{eqn:dpq}
\end{align}
$\Omega$ denotes the domain of integration of $\mathbf{x}$.
This Gaussian similarity measure has a high value if the spatial support of the two Gaussian mixtures aligns well.
It bears resemblance to the Bhattacharyya Coefficient~\cite{bhattacharyya_measure_1946} used to measure the similarity of probability distributions while being computationally less expensive.

\begin{figure}
  \centering
  \includegraphics[height=8cm]{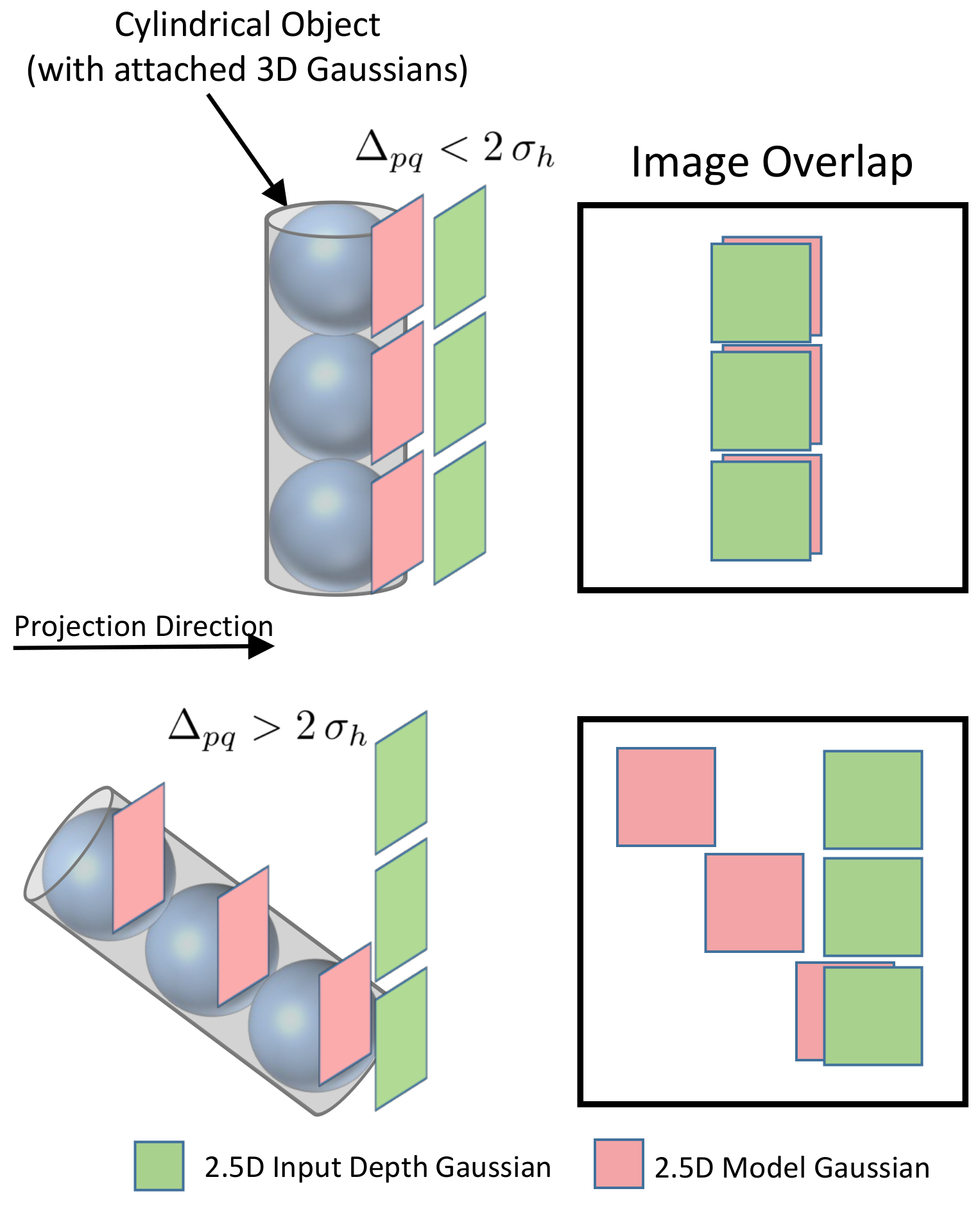}
  \caption{\textbf{Depth Similarity Term}: Consider the similarity value ($E_{sim}$) for a cylindrical shape represented by 3 Gaussians ($x \in \mathcal{R}^3$). The top figure shows a case where the value of $E_{sim}$ is high since the image overlap is high and the depth difference $\Delta_{pq}$ is low.
The bottom figure shows a case where the image overlap is moderate but $\Delta > 2 \, \sigma_h$ thus making $E_{sim} = 0$.}
  \label{fig:esim}
\end{figure}

\parahead{Depth Similarity Term ($E_{sim}$)}
The 2.5D depth similarity term measures the quality of overlap between the projected model Gaussian mixture $\C_p$ and the image Gaussian mixture $\C_I$.
Additionally, this measure also incorporates the depth information available for each Gaussian in the mixture.
Figure~\ref{fig:esim} explains this term intuitively.
Two Gaussians that are close (in 2D pixel distance) in the depth image obtain a high value if their depth values are also close.
On the other hand, the same Gaussians obtain a low value if their depths are too far apart.
Formally, this term is defined as,
\begin{align}
  E_{sim}(\C_p,\C_I)
  & = \frac{1}{E(\C_I, \C_I)} \, \sum_{p \in \C_p} \sum_{q \in \C_I} \Delta(p, q) \, D_{pq}
\end{align}
where $D_{pq}$ is as defined in Equation~\ref{eqn:dpq} and the \emph{depth similarity factor} is
\begin{align}
  \Delta(p, q) =
  \begin{cases}
    0, & \mbox{if } |d_p - d_q| \ge 2 \, \sigma_h\\ \label{eqn:dsf}
    1 - \frac{|d_p - d_q|}{2 \, \sigma_h}, & \mbox{if } |d_p - d_q| < 2 \, \sigma_h
  \end{cases}.
\end{align}
Here, $d_p$ and $d_q$ are the depth values associated with each Gaussian in $\C_p$ and $C_q$ respectively,
and $\sigma_h$ is the standard deviation of the \emph{unprojected} model Gaussian $\G_h$.
The surface depth value of each Gaussian in $C_p$ is computed as $d_p = \left[\MU_h\right]_z - \sigma_h$.
The factor ${E(\C_I, \C_I)}$ is the similarity measure from equation~\ref{eqn:gausssim} of the depth image with itself and serves to normalize the similarity term.
The $\Delta$ factor has a support $[0, 1]$ thus ensuring the similarity between a projected model Gaussian and an image Gaussian is $0$ if they lie too far apart in depth.

\parahead{Collision Penalty Term ($E_{col}$)}
The fingers of a hand are capable of fast motions and often come in close proximity with one another causing aliasing of corresponding depth pixels in the input.
Including a penalty for collisions avoids fingers \emph{sticking} with one another and Gaussian interpenetration.
The 3D Gaussian mixture representation of the hand model ($\C_h$) offers an efficient way to penalize collisions because they implicitly act as collision proxies.
We define the penalty for collisions as,
\begin{align}
  E_{col}(\Theta)
  & = \frac{1}{E(\C_h, \C_h)} \, \sum_{p \in \C_h} \sum_{\substack{q \in \C_h\\ q > p}} D_{pq},
\end{align}
where $E(\C_h, \C_h)$ is the similarity measure from equation~\ref{eqn:gausssim} for the hand model and serves to normalize the collision term.
The collision term penalizes model Gaussians that collide with others but not if they collide with themselves.
As we show in the results, the collision term has a large impact on tracking performance.

\parahead{Joint Limit Penalty Term ($\E_{lim}$)}
We add a penalty for poses that exceed predefined joint angle limits.
This forces biomechanically plausible poses to be preferred over other poses.
The joint limit penalty is given as,
\begin{align}
  E_{lim}(\Theta) = \sum_{\theta_j \in \Theta}
  \begin{cases}
    0, & \mbox{if } \theta_j^l \le \theta_j \le \theta_j^h\\
    ||\theta_j^l - \theta_j||^2, & \mbox{if } \theta_j < \theta_j^l\\
    ||\theta_j - \theta_j^h||^2, & \mbox{if } \theta_j > \theta_j^h\\
  \end{cases}
\end{align}
where $\theta_j^l$ and $\theta_j^h$ are the lower and higher limits of the parameter $\theta_j$ which is defined based on anatomical studies of the hand~\cite{simo_serra_kinematic_2011}.
The result is a tracked skeleton that looks biomechanically plausible.

\parahead{Smoothness Penalty Term ($\E_{smo}$)}
During frame-by-frame pose optimization, noise is introduced which manifests as jitter in tracking.
To prevent this we penalize fast motions by adding a penalty as done by~\cite{stoll_fast_2011}. This term is given as,
\begin{align}
  E_{smo}(\Theta) = \sum_{j = 0}^{|\Theta|-1} \left( 0.5 \, \left( \Theta_j^{t-2} + \Theta_j^t\right) - \Theta_j^{t-1}\right)^2
\end{align}
where, $\Theta^t$ denotes the pose at time $t$. This term acts as a regularizer and prevents jitter in the tracked pose.

\subsection{Detection-Guided Pose Optimization}\label{sec:detection_guided_pose}
%
%
To increase chances of recovery when the estimated pose is at a wrong local pose optima, we use a second pose optimization energy 
that includes evidence from hand part detection. 
In particular we use pixel labels computed with a trained random forest~\cite{criminisi2013decision}. 
Decision forests have been used before for 3D pose and joint position detection~\cite{keskin_real_2011,tang_latent_2014,tang_real-time_2013,xu_efficient_2013}.
We are interested in part labels and therefore follow an approach similar to \cite{shotton_real-time_2011} and \cite{keskin_real_2011}.
The evidence from the part labels is incorporated in our tracking.

We use 12 part labels for the hand (see Figure~\ref{fig:pipeline}) and found this to be an ideal trade-off between classification accuracy and sufficient evidence for detection-guided optimization.
We adopt the same depth features as \cite{shotton_real-time_2011}.
We use 50,000 labeled training images spanning the hand pose space.
As opposed to previous work~\cite{keskin_real_2011} that use  synthetic data, we use
\textbf{real} hand motions with part labels which were obtained using the depth-only version of our method and tracking motions slowly without causing tracking failure.
During training we trained 3 trees, each with a maximum depth of 22.
For each training image, we sampled 2000 random, foreground pixels, and evaluated 4000 candidate threshold-feature response pairs.

During quadtree clustering of the depth (Section~\ref{sec:depth_rep})
each quad is endowed with a part label, $l_q$ which is the label with the highest number of votes among all pixels in the quad.
We can now tightly integrate the part labels in the optimization by defining a pose fitting energy
identical to Equation~\ref{eqn:objective} with one exception: the depth similarity factor from Equation~\ref{eqn:dsf}
is replaced by the following \emph{label similarity factor}.
\begin{align}
  \Delta_l(p, q) =
  \begin{cases}
    0, & \mbox{if } l_p \neq l_q \mbox{ or } |d_p - d_q| \ge 2 \, R_i\\
    1 - \frac{|d_p - d_q|}{2 \, R_i}, & \mbox{if } l_p = l_q\nonumber
  \end{cases},
\end{align}
where $l_p$ and $l_q$ are the part labels
, $d_p$ and $d_q$ are the depth values
, $R_i$ refers to the radius of influence which is set to 200~mm in all our experiments.
Intuitively, $\Delta_l$ has a value of zero if the labels are different
and a value of one if the two Gaussians have identical labels and are perfectly aligned in 2.5D.
The labels $l_p$ are obtained from preassigned labels of each Gaussian in the hand model.

\section{Late Fusion}\label{sec:fusion}
%
The goal of optimization is to find the pose $\Theta$ such that $-\mathcal{E}(\Theta)$ is minimized.
Our energies---both with and without detection---are well suited for gradient based optimization because we can derive the analytic gradient with
respect to the DOFs $\Theta$.
For efficiency, we adopt the fast gradient-based optimizer with adaptive step length proposed by~\cite{stoll_fast_2011}.

To improve robustness, especially with changes in direction and global rotation, we use multiple \emph{pose particles} for optimizing each frame.
Multiple particles improve the chances of a good initialization for optimization.
Each particle $P_i$ is initialized using the pose parameters from two previous
time steps $\Theta^{t-1}$ and $\Theta^{t-2}$ with different extrapolation factors $\alpha_{ij}$ for each DOF $j$.
This is given as
$P_i = \theta^{t-1}_j + \alpha_{ij} \, \theta^{t-2}_j, \, \forall j$.
We sample $\alpha_{ij}$ from a normal distribution with mean fixed at the initial value of $\theta_j$.
All but one of these particles is optimized using the depth-only energy.
Finally, the pose particle which converges with the best energy value is chosen
as the winning pose. In all our experiments, we found that 2--3 particles
were sufficient to obtain more robust results. Increasing the particles
had a negative effect and caused jitter in the final pose.
Each particle used 10--30 iterations per frame.
We justify the choice of these parameters in Section~\ref{sec:results}.
\section{User Specific Hand Modeling}\label{sec:handmodeling}
Accounting for the fact that there are large variations in anthropometric dimensions, our pose optimization method works best with a customized hand model for each user.
Our method does not necessitate laser scans, manual tuning of the model, nor semi-automatic bone model optimization as used by existing methods~\cite{stoll_fast_2011,sridhar_interactive_2013}.

\begin{figure}
  \centering
  \subfigure{\includegraphics[height=0.15\textwidth]{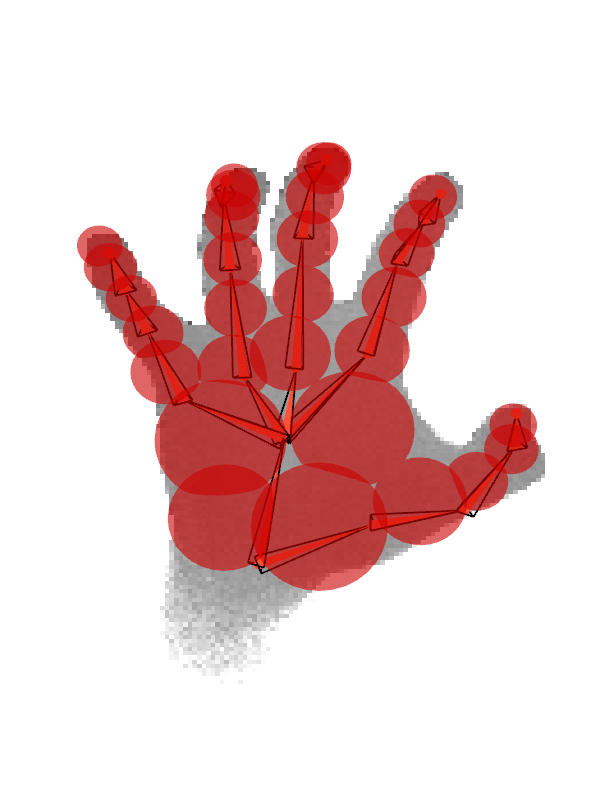}}
  \subfigure{\includegraphics[height=0.15\textwidth]{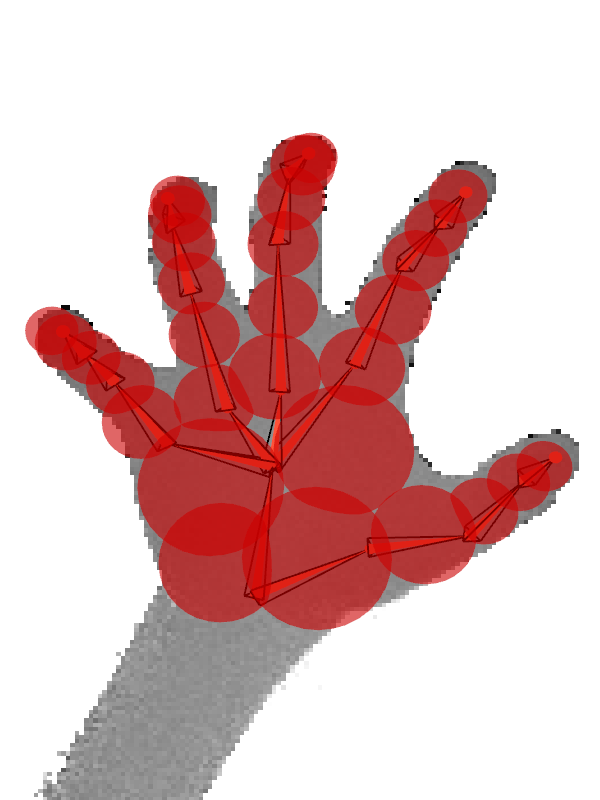}}
  \subfigure{\includegraphics[height=0.15\textwidth]{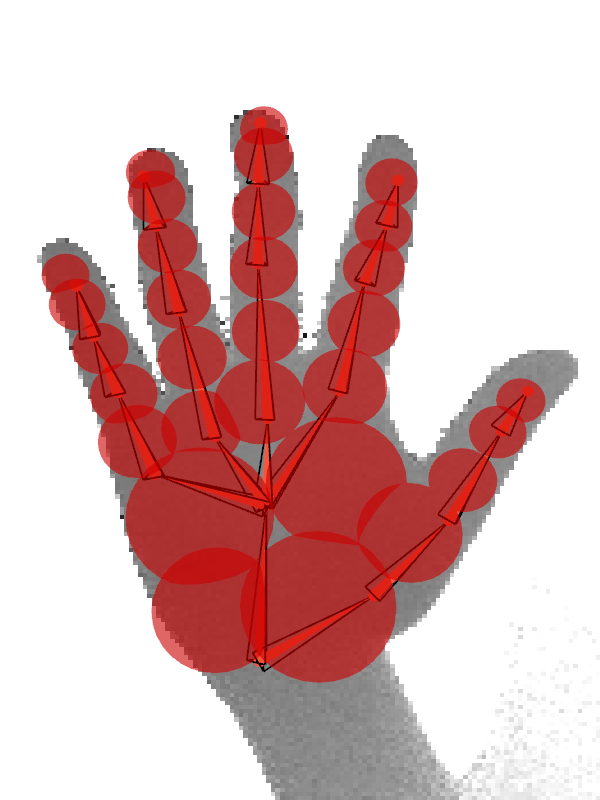}}
  \subfigure{\includegraphics[height=0.15\textwidth]{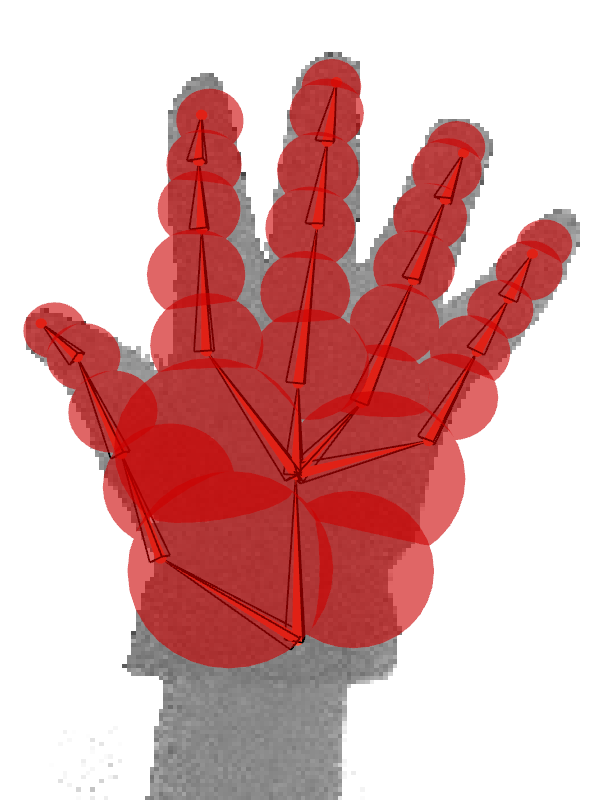}}
  \caption{Automatic fitting of user specific hand model for 4 subjects, one of whom is wearing a thick glove to simulate variability in hand dimension. The red spheres denote 3D Gaussians.}
  \label{fig:modelfit}
\end{figure}

We observed in our experiments that the primary variations in hand dimensions are finger thickness, hand length and width.
We developed a simple strategy where a default hand model is scaled using three parameters: hand length, width, and variance of Gaussians.
To find the scaling parameters for a user, we perform a greedy search over a fixed range for each scaling parameter.
At each point on this parameter grid we evaluate the energy function value from Equation~\ref{eqn:objective}.
The parameters that obtain the best energy are selected as the model scaling parameters.
This method is fast and takes less than a second to find a user-specific hand model.
Figure~\ref{fig:modelfit} shows some qualitative results from our model fitting strategy for different users.

\section{Results and Evaluation}\label{sec:results}
%
%
%
We provide quantitative and qualitative evidence for performance with fast motions and finger articulations.
Evaluation of hand tracking algorithms is challenging because ground truth data is difficult to obtain.
Marker-based motion capture is often problematic due to self-occlusions.
Many methods have therefore resorted to evaluation on synthetic data~\cite{oikonomidis_efficient_2011,oikonomidis_full_2011} which, however, is not representative of real hand motions.
There are also no established benchmark datasets with accepted error metrics, and only a few implementations have been made public.

We use the dataset from \cite{sridhar_interactive_2013} which consists of seven challenging sequences (abduction--adduction, finger counting, finger waving, flexion--extension, pinching, random motions, grasping)
that are further split into slow and fast parts.
%
The fingertips are annotated manually in the depth data thus making it possible to compare with the multi-view approaches of \cite{sridhar_interactive_2013} and \cite{sridhar2014real}.
Additionally, we also compare with the discriminative method of \cite{tang_latent_2014} on 3 sequences.
We also motivate the need for our fusion strategy, parameter selection in optimization, and analyze the effects of different components of our objective function.
We also provide details about our framerate and qualitative evidence of improvements over \cite{melax_dynamics_2013} and the Leap Motion.
Please see the supplementary material for more results.

\parahead{Error Metrics}
Our evaluations concern the average fingertip localization error which correlates well with overall pose accuracy.
For each sequence, we compute Euclidean error of the 5 fingertip positions averaged over all frames.
Additionally, we use a second error metric~\cite{qian_realtime_2014} which is the percentage of frames that have an error of less than $x$~mm where $x \in \{15, 20, 25, 30\}$.
This is a stricter measure that highlights reliability.

\subsection{Quantitative Evaluation}
\begin{figure}
  \centering
  \includegraphics[width=\columnwidth]{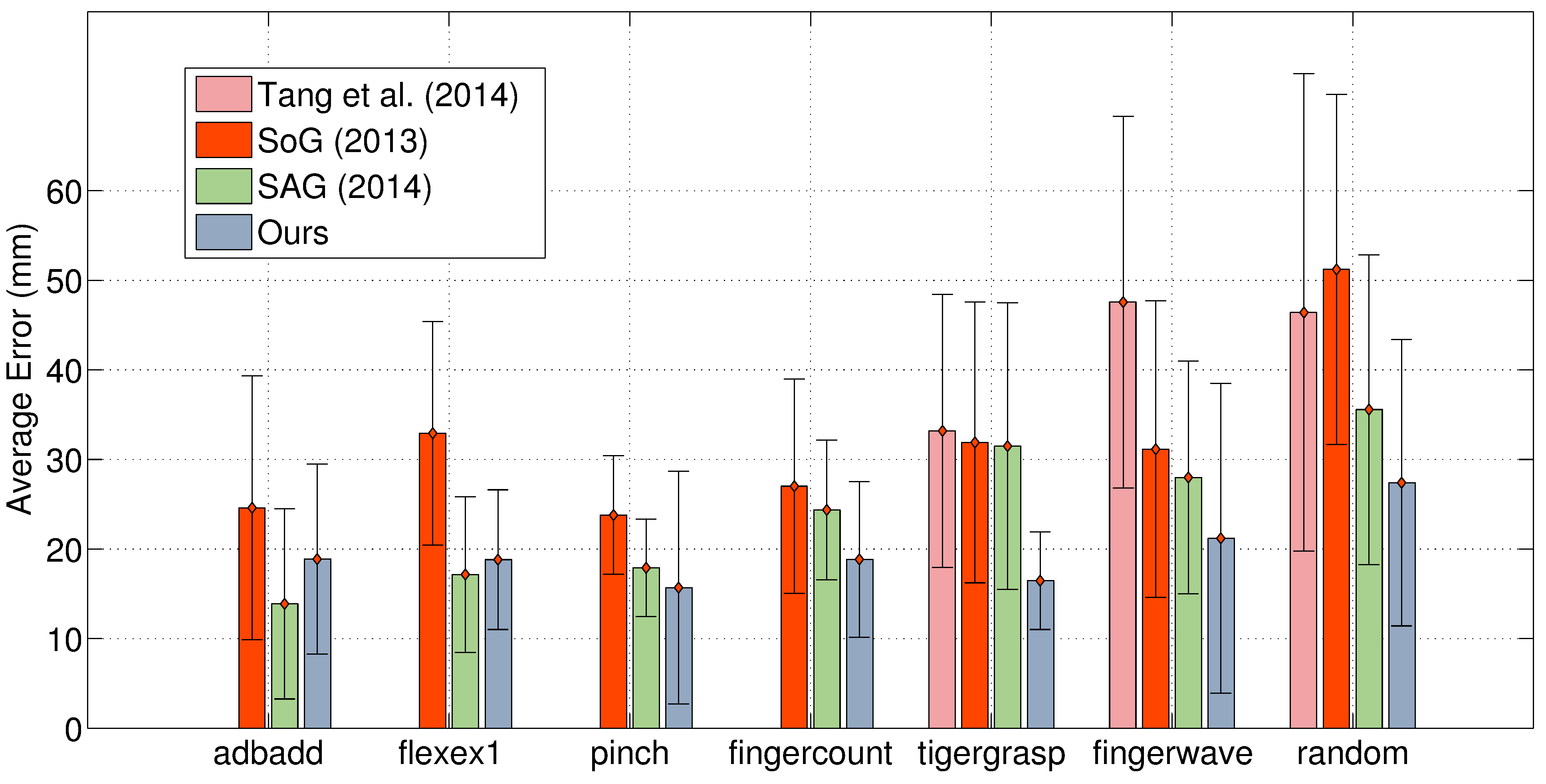}
  \caption{Average error over the 7 sequences in Dexter 1 and comparison with the \textbf{multi-view} methods of \cite{sridhar_interactive_2013} and \cite{sridhar2014real}, and the \textbf{detection-based} method of \cite{tang_latent_2014}. Our method achieves the lowest error on 5 of the 7 sequences and the best average error ($\textbf{19.6}$~mm).}
  \label{fig:dexter1}
\end{figure}

\parahead{Accuracy}
Figure~\ref{fig:dexter1} shows our average error compared with that of \cite{sridhar_interactive_2013}, \cite{sridhar2014real}, and \cite{tang_latent_2014}.
Our method produces the lowest average error of $\textbf{19.6}$~mm while using only a single depth camera.
The multi-view approaches of \cite{sridhar2014real} and \cite{sridhar_interactive_2013} have errors of $\textbf{24.1}$~mm and $\textbf{31.8}$~mm respectively.
The detection-based discriminative method of \cite{tang_latent_2014} has an error of $\textbf{42.4}$~mm (3 sequences only) highlighting the need for using temporal information.
%
We observe that our method does particularly well for motions that involve articulation of fingers such as \texttt{flexex1}.
Our worst performance was on the \texttt{random} sequence involving fast global hand rotation.
%

\parahead{Error Frequency}
Table 1 confirms the trend that our method performs well for finger articulations.
In 6 out of 7 sequences, our method results in tracking errors of less than 30~mm in 85\% of the frames.
A closer examination shows that these sequences contain complex finger articulations.
%
\begin{table}
  \centering
  \begin{tabular}{|c||c|c|c|c|}
    \hline
    \tabhead{Error $<$ (mm)} & \tabhead{adbadd} & \tabhead{fingercount} & \tabhead{fingerwave}\\
    \hline
    15 & 56.6 & 50.0 & 56.2\\
    20 & 70.6 & 66.5 & 71.2 \\
    25 & 76.2 & 77.7 & 78.3 \\
    30 & 84.9 & 85.8 & 85.0 \\
    \hline
  \end{tabular}

  \begin{tabular}{|>{\centering\arraybackslash}m{1.57cm}||c|c|c|c|}

    \hline
    \tabhead{Error $<$ (mm)} & \tabhead{flexex1} & \tabhead{pinch} & \tabhead{random} & \tabhead{tigergrasp}\\
    \hline
    15 &  53.7 & 56.7 & 19.1 & 62.9\\
    20 &  68.1 & 83.9 & 40.7 & 80.6 \\
    25 &  76.7 & 93.1 & 59.0 & 87.3 \\
    30 &  85.5 & 97.4 & 70.6 & 91.8\\
    \hline
  \end{tabular}
  \caption{Percentage of total frames in a sequence that have an error of less $x$~mm.}
\end{table}

\begin{figure*}
  \centering
  \includegraphics[width=\textwidth]{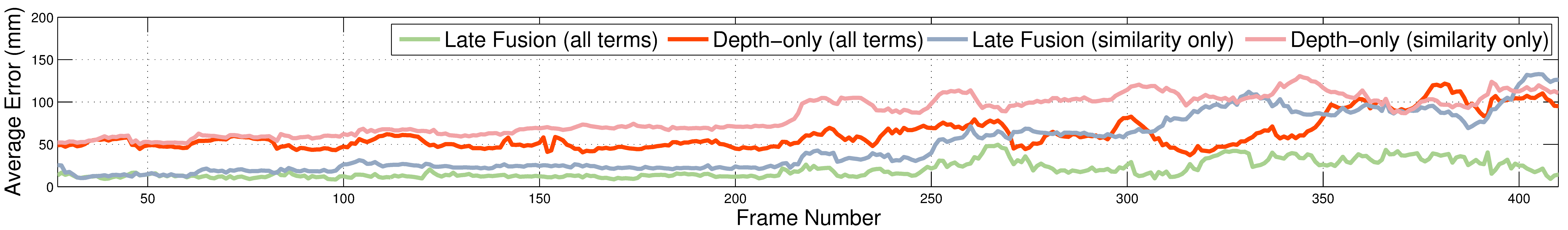}
  \vspace{-4mm}
  \caption{Plot of the error for the depth-only tracking and late fusion approach. Each approach was run with only the similarity term $E_{sim}$ and with all terms. Notice the catastrophic tracking failure with the depth-only energy. The late fusion strategy is robust and prevents error accumulation. The collision penalty term also results in large accuracy gains. Best viewed in color.}
  \label{fig:flexex}
  \vspace{-0.5cm}
\end{figure*}

\parahead{Robustness}
We measure robustness as the ability of a tracker to recover from tracking failures.
To demonstrate how our late fusion strategy and the different terms in the energy help achieve this, we show the frame-wise error
over the \texttt{flexex1} sequence (Figure~\ref{fig:flexex}).
Using the depth-only energy with all terms except $E_{sim}$ disabled (2 particles) results
in catastrophic tracking failure as shown by the accumulating error.
Adding the other terms, especially the collision penalty ($E_{col}$) term,  improves accuracy but results are still unsatisfactory.
The results from the late fusion approach show large gains in accuracy.
The errors also remain more uniform which results in temporally stable tracking with less jitter.
%
\begin{figure}
  \centering
   \subfigure{\includegraphics[height=0.15\textwidth]{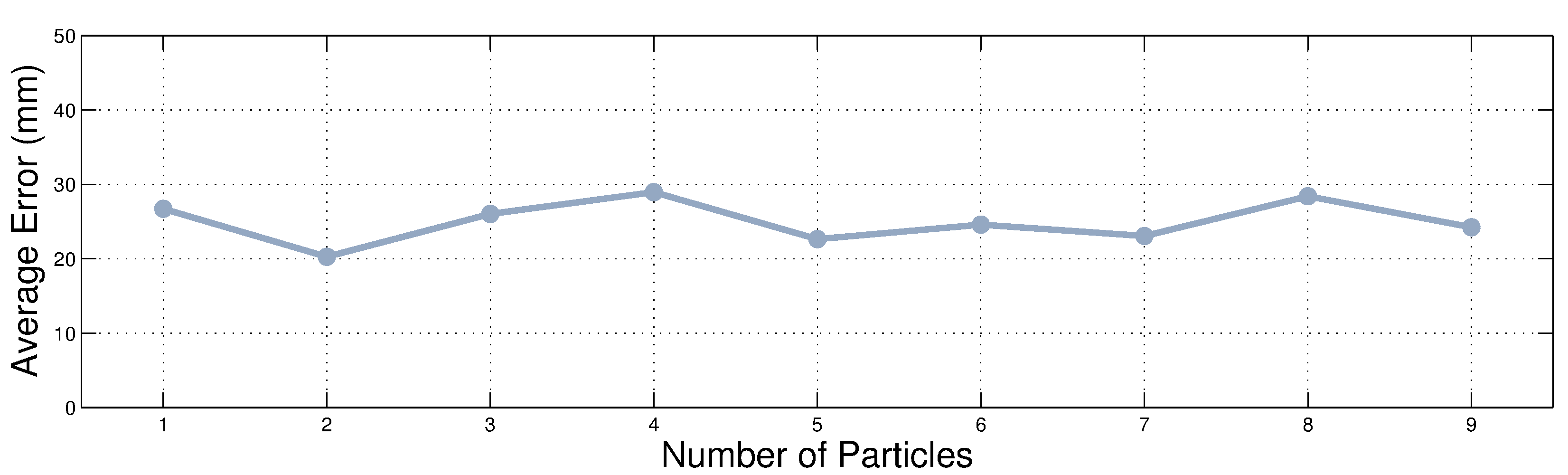}}
   \subfigure{\includegraphics[height=0.15\textwidth]{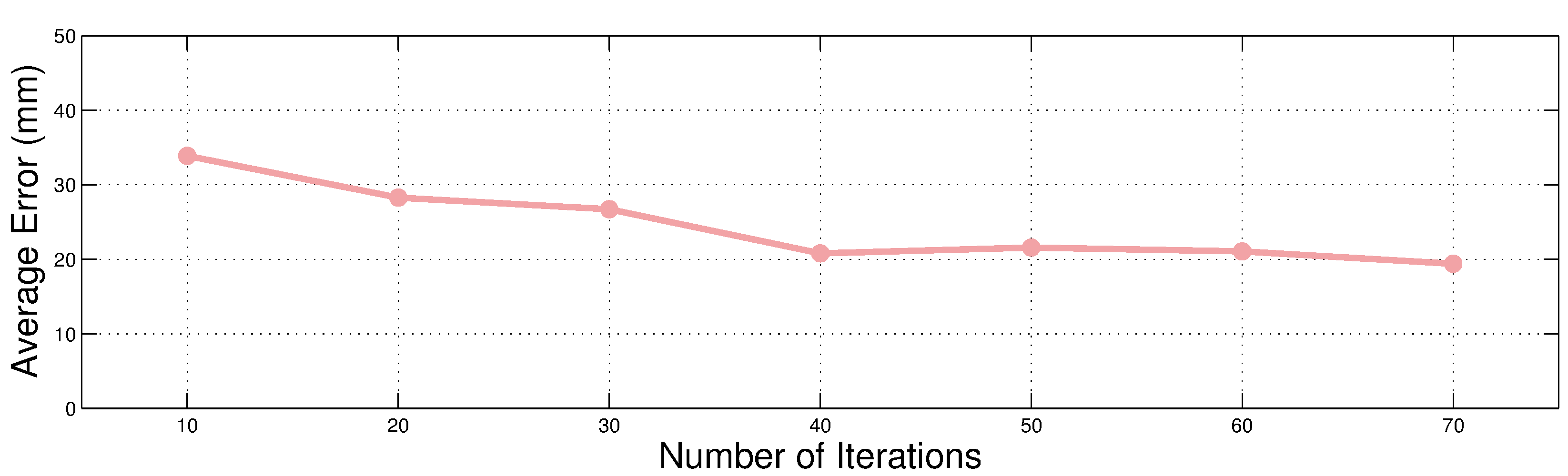}}
   \caption{Effect of varying the number of particles and iterations during optimization. We found that increasing the number of particles resulted in diminishing returns.}
   \label{fig:iter}
\end{figure}

\parahead{Number of Particles and Iterations}
Figure~\ref{fig:iter} shows the effect of varying the number of particles and iterations during optimization.
As the number of particles increased we noticed very little increase in accuracy. In fact, the best accuracy was with 2 particles which we use throughout.
We noticed a reduction in error when using more number of iterations per particle but at the expense of runtime.
We therefore fixed the number of iterations to be 10.
\begin{figure*}[!htb]
  \centering
   \subfigure{\includegraphics[height=0.15\textwidth]{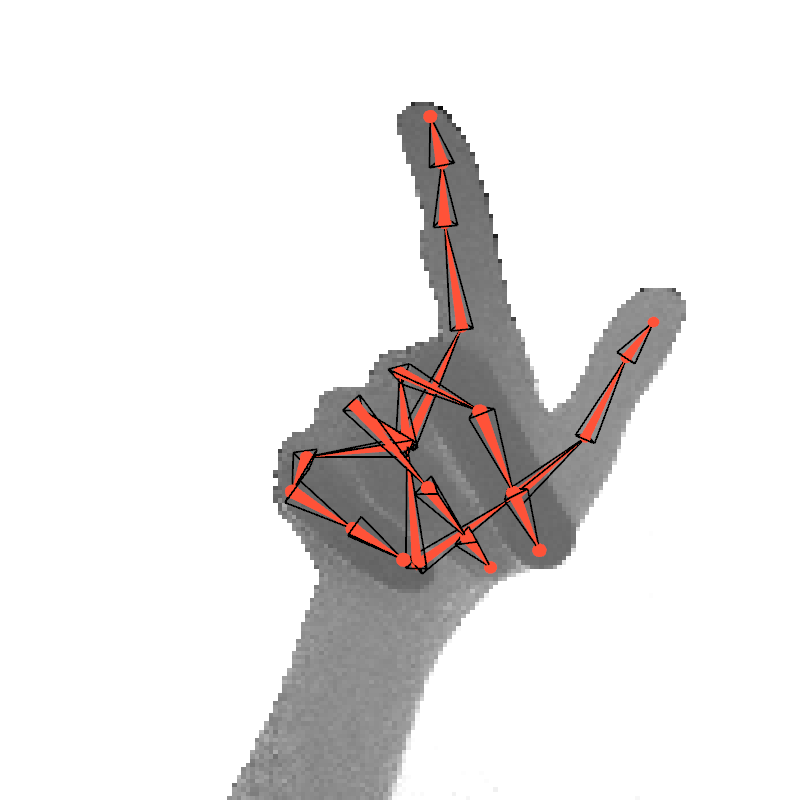}}
   \subfigure{\includegraphics[height=0.15\textwidth]{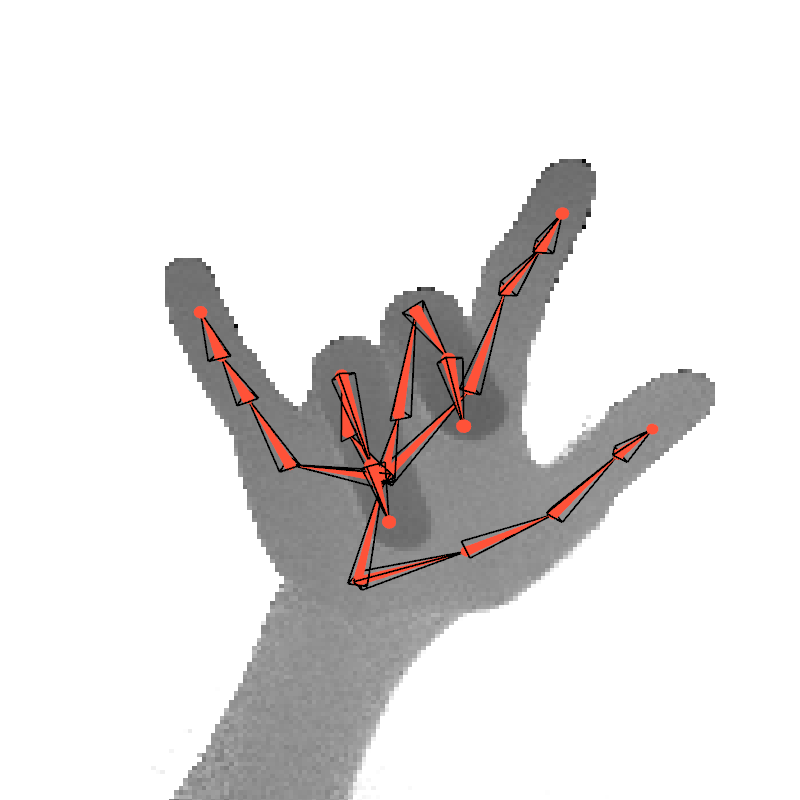}}
   \subfigure{\includegraphics[height=0.15\textwidth]{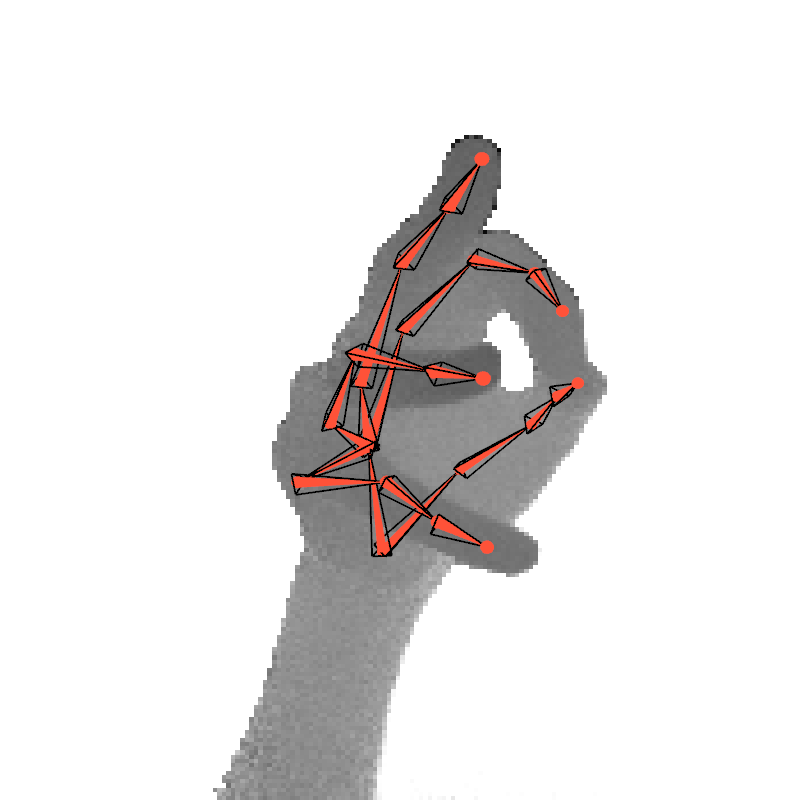}}
   \subfigure{\includegraphics[height=0.15\textwidth]{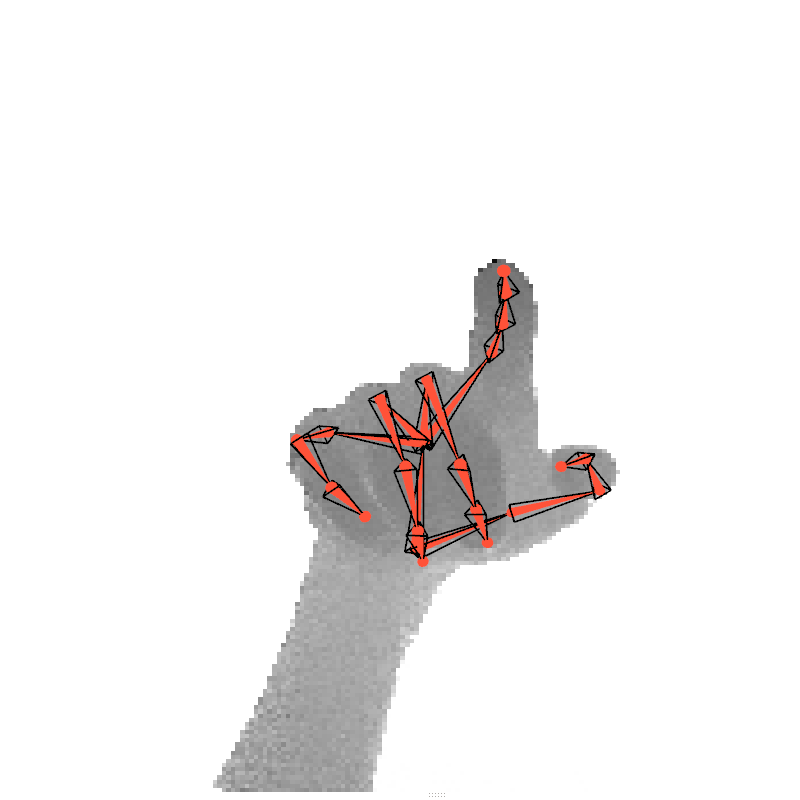}}
   \subfigure{\includegraphics[height=0.15\textwidth]{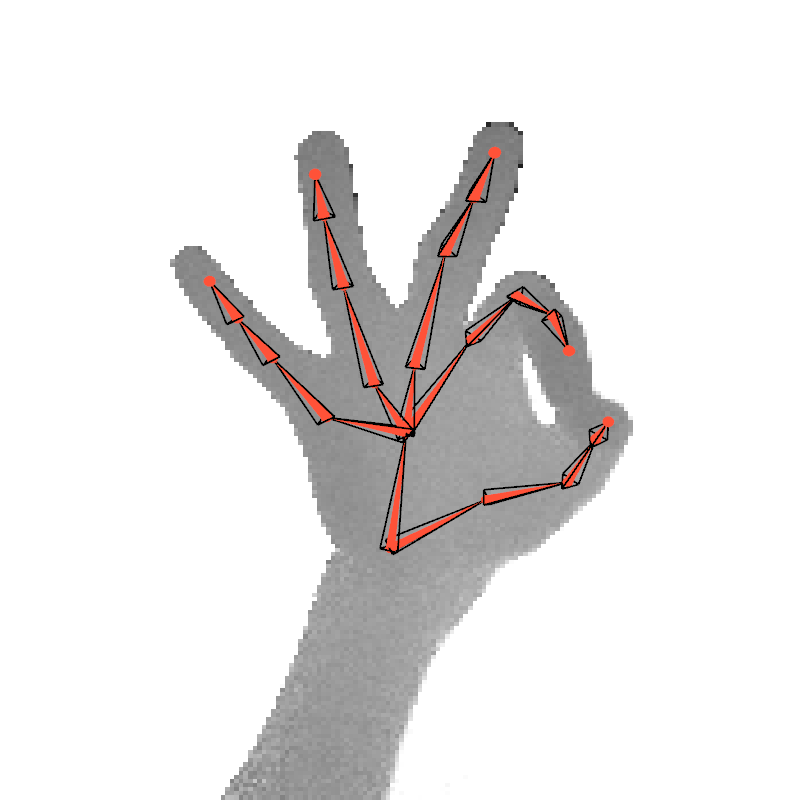}}
   \subfigure{\includegraphics[height=0.15\textwidth]{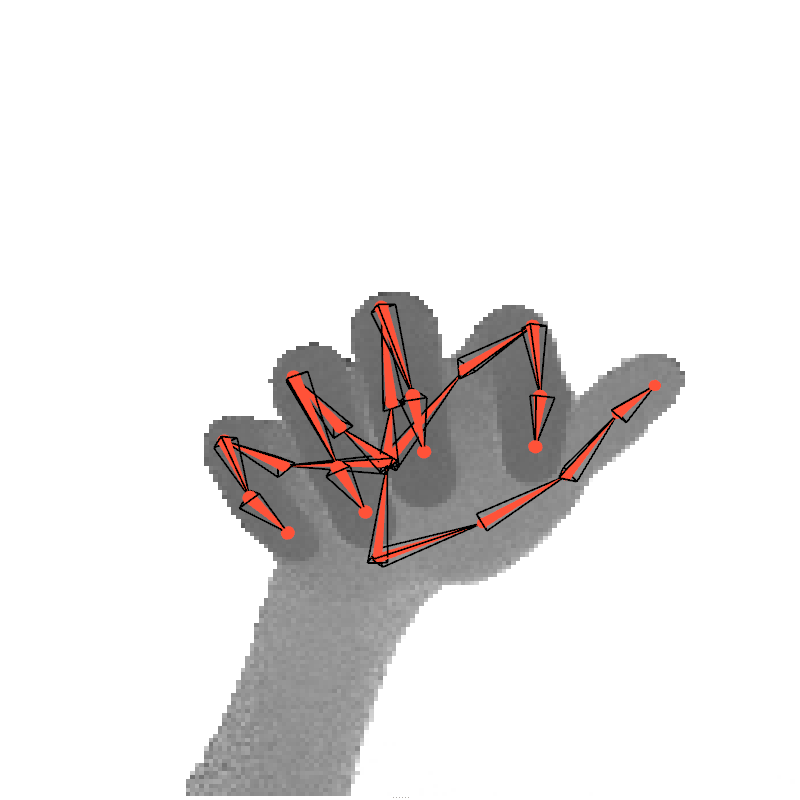}}
   \subfigure{\includegraphics[height=0.15\textwidth]{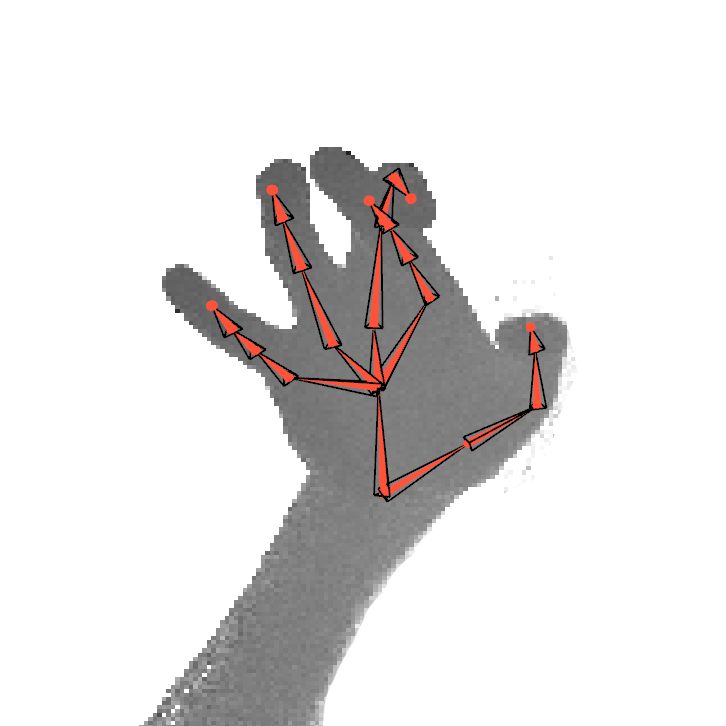}}
   \subfigure{\includegraphics[height=0.15\textwidth]{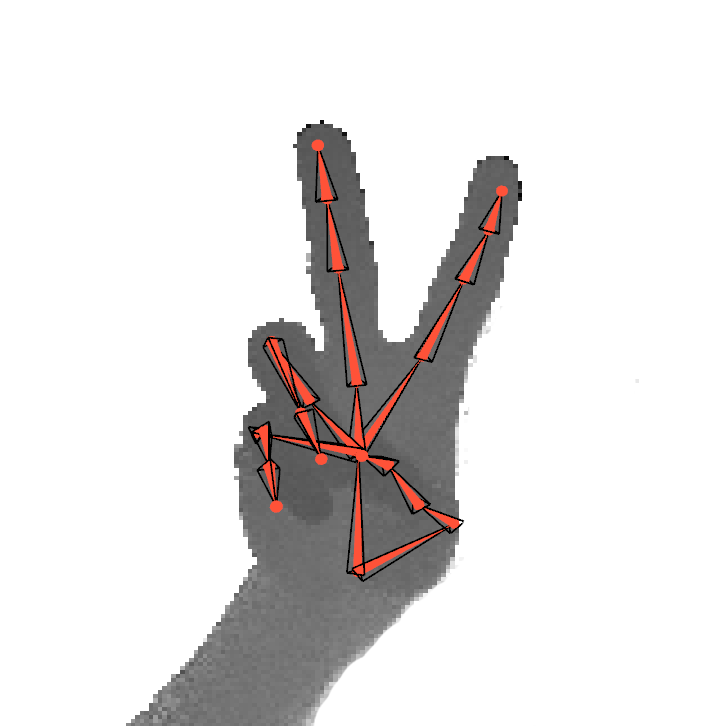}}
   \subfigure{\includegraphics[height=0.15\textwidth]{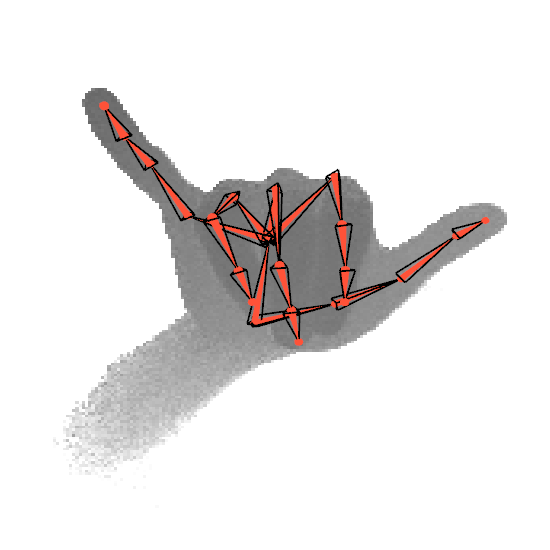}}
   \subfigure{\includegraphics[height=0.15\textwidth]{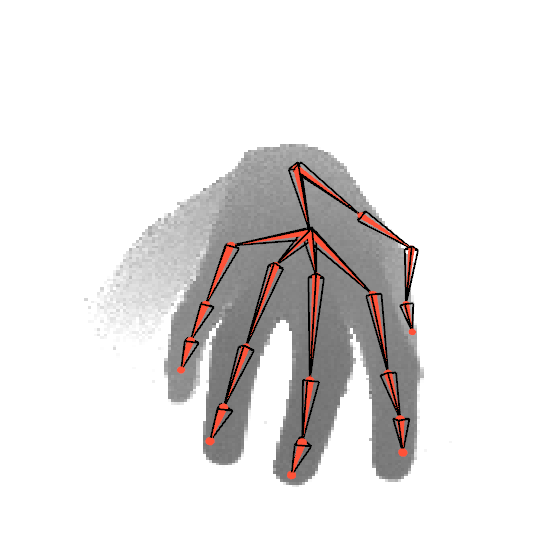}}
\framebox{
   \subfigure{\includegraphics[height=0.15\textwidth]{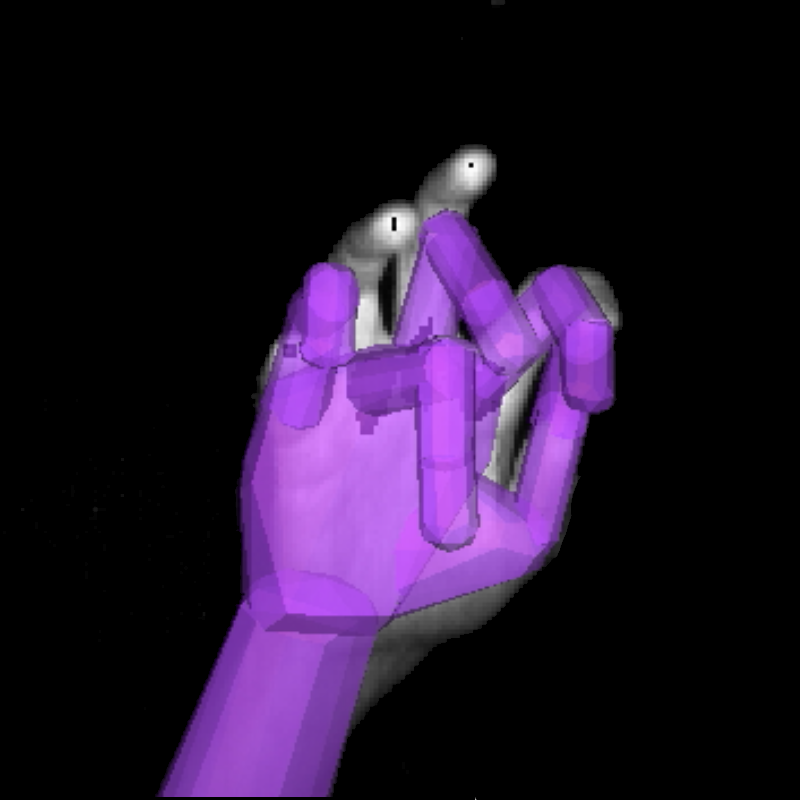}}
   \subfigure{\includegraphics[height=0.15\textwidth]{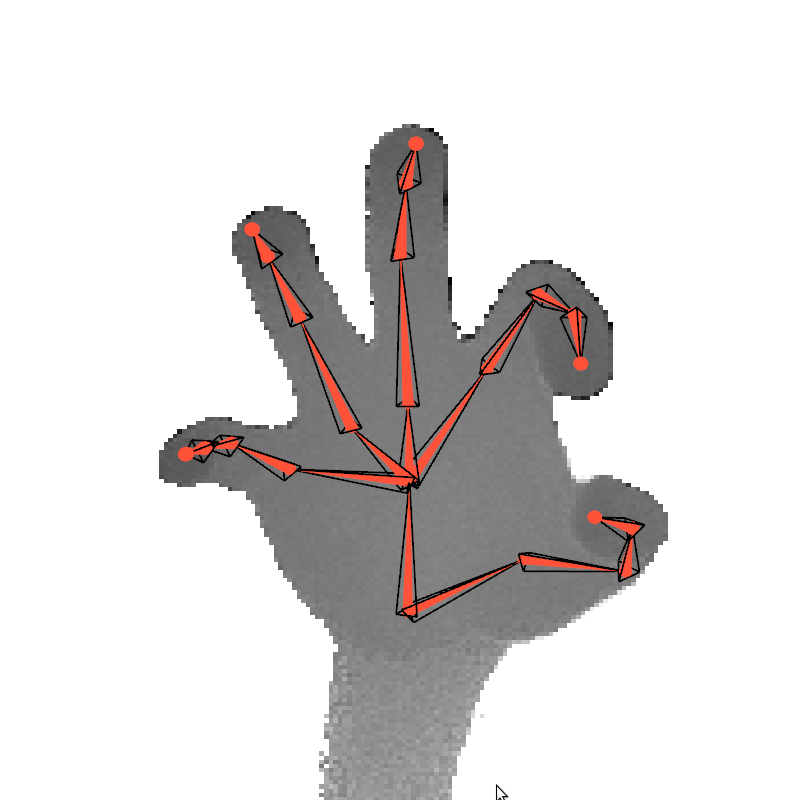}}
}
\framebox{
   \subfigure{\includegraphics[height=0.15\textwidth]{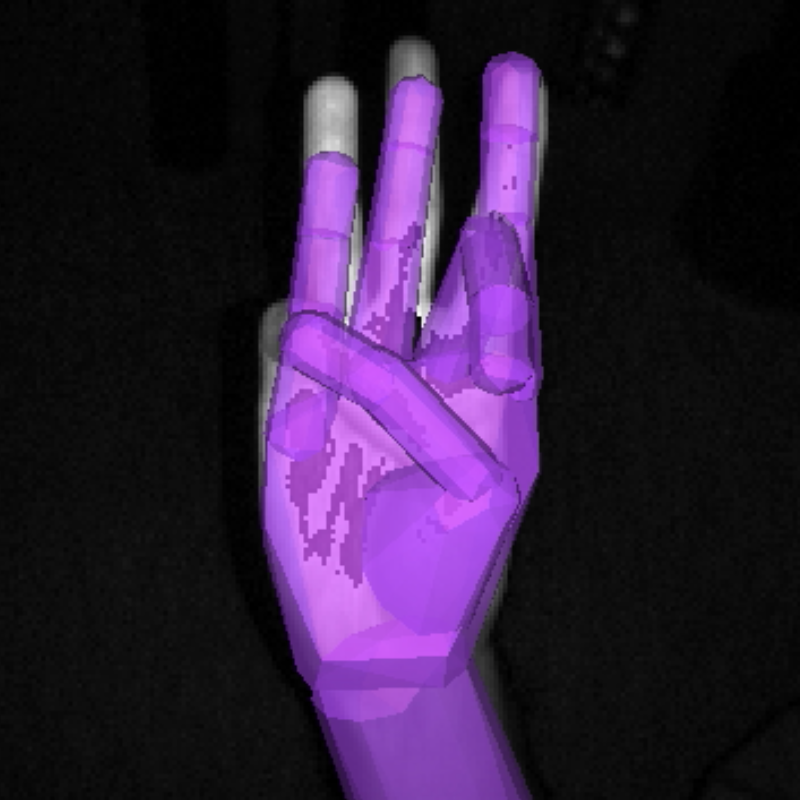}}
   \subfigure{\includegraphics[height=0.15\textwidth]{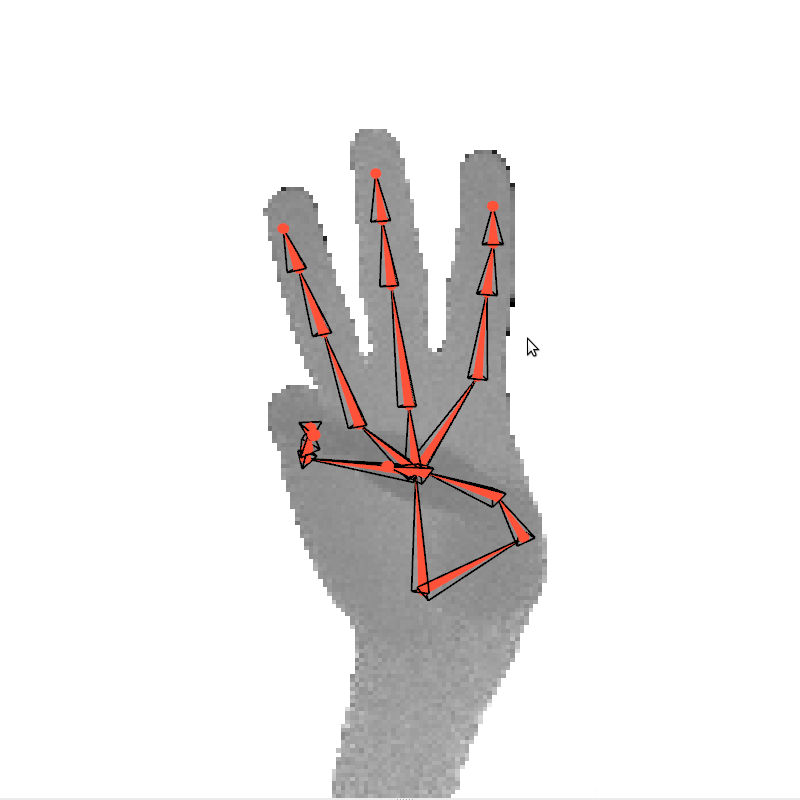}}
}
\framebox{
   \subfigure{\includegraphics[height=0.15\textwidth]{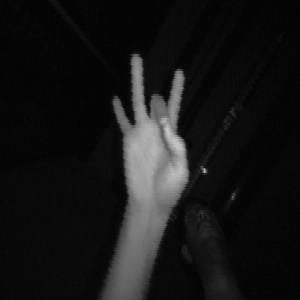}}
   \subfigure{\includegraphics[height=0.15\textwidth]{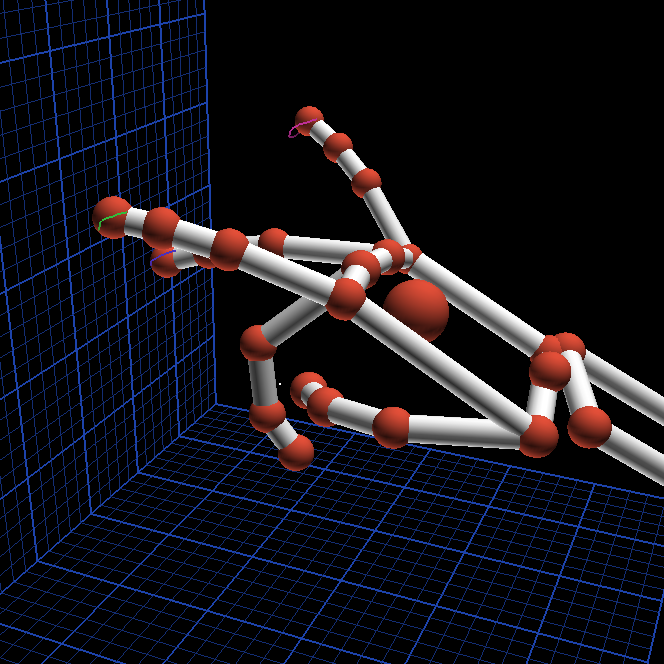}}
   \subfigure{\includegraphics[height=0.15\textwidth]{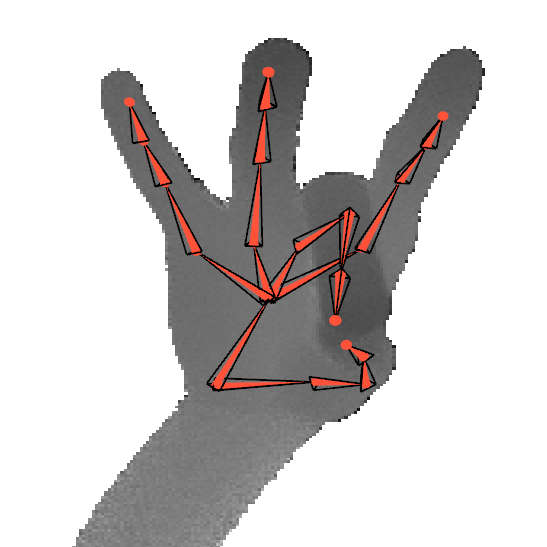}}
   \subfigure{\includegraphics[height=0.15\textwidth]{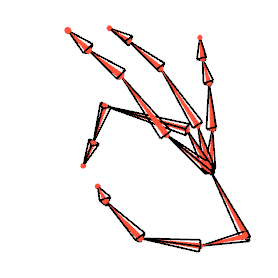}}
}
\vspace{2mm}
  \caption{Qualitative results from our tracking approach (top row and four leftmost in the second row). The highlighted boxes show comparison with \cite{melax_dynamics_2013} and the Leap Motion both of which produce a finger sliding effect. Our method tracks the pinch faithfully.}
  \label{fig:quals}
\end{figure*}


\parahead{Tracking Speed}
We tested the tracking speed of different variants of our method on a 3.6~GHz Intel Processor with 16~GB of RAM.
Our method was parallelized using OpenMP but no GPU was used.
All tests were done with the Intel Senz3D depth camera with a depth resolution of 320 $\times$ 240 and capture rate of 60~fps.
The decision forest when loaded in memory used 1~GB because the trees were stored as full trees.
This can be avoided by loading only nodes that are valid.
The depth-only energy when used with 1 particle, and 10 iterations per particle ran at 120~fps.
When 2 particles were used, the speed came down to 60~fps.
The late fusion approach, when used with 2 particles (10 iterations per particle), achieved a framerate of \textbf{50}~fps.
Image acquisition, part labeling, preprocessing, and creating the Gaussian mixture representation took 2~ms.
The optimization took between 18 and 20~ms.

\subsection{Qualitative Results}
We present several qualitative results from realtime sequences in Figure~\ref{fig:quals}.
The examples show motions with a wide range of finger articulations involving
abduction, adduction, flexion, extension, and considerable occlusions.
They also include common gestures such as the \emph{v-sign} and pointing.
In the boxes, we also show comparison with \cite{melax_dynamics_2013} and the Leap Motion on similar poses.
We observe a finger sliding effect in both these methods.
Pinching is an important gesture, but the Leap Motion produces sliding fingertips
which makes it hard to detect pinching gestures from the tracked skeleton.
Our method reproduces pinching faithfully as is evident from the skeleton overlaid on the depth image.
Occasional tracking failures occur with large global rotations but the detection-guided energy eventually reinitializes to the correct pose.
Please see the supplementary materials for more results on different camera arrangements, failure cases, and tracking with different subjects.

\section{Conclusion}
In this paper, we presented a method for realtime hand tracking using detection-guided optimization.
Our method is robust and tracks the hand at 50~fps without using a GPU.
We contribute to the tracking literature by proposing a novel representation of the input data and hand model using a mixture of Gaussians.
This representation allows us to formulate pose estimation as an optimization problem and efficiently optimize it using analytic gradient.
We also showed how additional evidence from part detection can be incorporated into our tracking framework to increase robustness.
We evaluated our method on a publicly available dataset and compared with other state-of-the-art methods.
%
An important direction for future work is the tracking of multiple hands interacting with each other or with objects.
We believe that the strong analytic formulation offered by our method can help solve this.
%

\parahead{Acknowledgments}
This research was funded by the ERC Starting Grant projects CapReal (335545) and COMPUTED (637991), and the Academy of Finland.
We would like to thank Christian Richardt.

{\small
\bibliographystyle{ieee}
\bibliography{content/CVPR2015}

\begin{thebibliography}{10}\itemsep=-1pt

\bibitem{athitsos_estimating_2003}
V.~Athitsos and S.~Sclaroff.
\newblock Estimating {3D} hand pose from a cluttered image.
\newblock In {\em Proc. of CVPR 2003}, pages II--432--9 vol.2.

\bibitem{baak_data-driven_2011}
A.~Baak, M.~Muller, G.~Bharaj, H.-P. Seidel, and C.~Theobalt.
\newblock A data-driven approach for real-time full body pose reconstruction
  from a depth camera.
\newblock In {\em Proc. of ICCV 2011}, pages 1092--1099.

\bibitem{ballan_motion_2012}
L.~Ballan, A.~Taneja, J.~Gall, L.~Van~Gool, and M.~Pollefeys.
\newblock Motion capture of hands in action using discriminative salient
  points.
\newblock In {\em Proc. of {ECCV} 2012}, volume 7577 of {\em LNCS}, pages
  640--653.

\bibitem{bhattacharyya_measure_1946}
A.~Bhattacharyya.
\newblock On a measure of divergence between two multinomial populations.
\newblock {\em Sankhya: The Indian Journal of Statistics (1933-1960)},
  7(4):401--406, July 1946.

\bibitem{criminisi2013decision}
A.~Criminisi and J.~Shotton.
\newblock {\em Decision forests for computer vision and medical image
  analysis}.
\newblock Springer, 2013.

\bibitem{Fanello_2014}
S.~R. Fanello, C.~Keskin, S.~Izadi, P.~Kohli, D.~Kim, D.~Sweeney, A.~Criminisi,
  J.~Shotton, S.~B. Kang, and T.~Paek.
\newblock Learning to be a depth camera for close-range human capture and
  interaction.
\newblock {\em ACM TOG}, 33(4):86:1--86:11.

\bibitem{hutchison_real-time_2012}
V.~Ganapathi, C.~Plagemann, D.~Koller, and S.~Thrun.
\newblock Real-time human pose tracking from range data.
\newblock In {\em Proc. of {ECCV} 2012}, volume 7577 of {\em LNCS}, pages
  738--751.

\bibitem{girshick_efficient_2011}
R.~Girshick, J.~Shotton, P.~Kohli, A.~Criminisi, and A.~Fitzgibbon.
\newblock Efficient regression of general-activity human poses from depth
  images.
\newblock In {\em Proc. of ICCV 2011}, pages 415--422.

\bibitem{hamer_tracking_2009}
H.~Hamer, K.~Schindler, E.~Koller-Meier, and L.~Van~Gool.
\newblock Tracking a hand manipulating an object.
\newblock In {\em Proc. of ICCV 2009}, pages 1475--1482.

\bibitem{keskin_real_2011}
C.~Keskin, F.~Kirac, Y.~Kara, and L.~Akarun.
\newblock Real time hand pose estimation using depth sensors.
\newblock In {\em Proc. of ICCV Workshops 2011}, pages 1228--1234.

\bibitem{kim_digits:_2012}
D.~Kim, O.~Hilliges, S.~Izadi, A.~D. Butler, J.~Chen, I.~Oikonomidis, and
  P.~Olivier.
\newblock Digits: freehand {3D} interactions anywhere using a wrist-worn
  gloveless sensor.
\newblock In {\em Proc. of UIST 2012}, pages 167--176.

\bibitem{kurmankhojayev_monocular_2013}
D.~Kurmankhojayev, N.~Hasler, and C.~Theobalt.
\newblock Monocular pose capture with a depth camera using a sums-of-gaussians
  body model.
\newblock In {\em Pattern Recognition}, number 8142 in LNCS, pages 415--424.
  Jan. 2013.

\bibitem{lee_spacetop:_2013}
J.~Lee, A.~Olwal, H.~Ishii, and C.~Boulanger.
\newblock {SpaceTop:} integrating {2D} and spatial {3D} interactions in a
  see-through desktop environment.
\newblock In {\em Proc. of CHI 2013}, pages 189--192.

\bibitem{melax_dynamics_2013}
S.~Melax, L.~Keselman, and S.~Orsten.
\newblock Dynamics based {3D} skeletal hand tracking.
\newblock In {\em Proc. of I3D 2013}, pages 184--184.

\bibitem{oikonomidis_efficient_2011}
I.~Oikonomidis, N.~Kyriazis, and A.~Argyros.
\newblock Efficient model-based {3D} tracking of hand articulations using
  kinect.
\newblock In {\em Proc. of BMVC 2011}, pages 101.1--101.11.

\bibitem{oikonomidis_full_2011}
I.~Oikonomidis, N.~Kyriazis, and A.~Argyros.
\newblock Full {DOF} tracking of a hand interacting with an object by modeling
  occlusions and physical constraints.
\newblock In {\em Proc. of {ICCV} 2011}, pages 2088--2095.

\bibitem{Oikonomidis:CVPR2012}
I.~Oikonomidis, N.~Kyriazis, and A.~Argyros.
\newblock Tracking the articulated motion of two strongly interacting hands.
\newblock In {\em Proc. of CVPR 2012}, pages 1862--1869, June 2012.

\bibitem{oikonomidis_evolutionary_2014}
I.~Oikonomidis, M.~Lourakis, and A.~Argyros.
\newblock Evolutionary quasi-random search for hand articulations tracking.
\newblock In {\em Proc. of CVPR 2014}, pages 3422--3429.

\bibitem{qian_realtime_2014}
C.~Qian, X.~Sun, Y.~Wei, X.~Tang, and J.~Sun.
\newblock Realtime and robust hand tracking from depth.
\newblock In {\em Proc. of CVPR 2014}, pages 1106--1113.

\bibitem{shotton_real-time_2011}
J.~Shotton, A.~Fitzgibbon, M.~Cook, T.~Sharp, M.~Finocchio, R.~Moore,
  A.~Kipman, and A.~Blake.
\newblock Real-time human pose recognition in parts from single depth images.
\newblock In {\em Proc. of CVPR 2011}, pages 1297--1304.

\bibitem{simo_serra_kinematic_2011}
E.~Simo~Serra.
\newblock {\em Kinematic Model of the Hand using Computer Vision}.
\newblock PhD thesis, Institut de Rob\`otica i Inform\`atica Industrial, 2011.

\bibitem{sridhar_interactive_2013}
S.~Sridhar, A.~Oulasvirta, and C.~Theobalt.
\newblock Interactive markerless articulated hand motion tracking using {RGB}
  and depth data.
\newblock In {\em Proc. of ICCV 2013}, pages 2456--2463.

\bibitem{sridhar2014real}
S.~Sridhar, H.~Rhodin, H.-P. Seidel, A.~Oulasvirta, and C.~Theobalt.
\newblock Real-time hand tracking using a sum of anisotropic gaussians model.
\newblock In {\em Proc. of 3DV 2014}, pages 319--326.

\bibitem{stoll_fast_2011}
C.~Stoll, N.~Hasler, J.~Gall, H.~Seidel, and C.~Theobalt.
\newblock Fast articulated motion tracking using a sums of gaussians body
  model.
\newblock In {\em Proc. of ICCV 2011}, pages 951--958.

\bibitem{sturman_survey_1994}
D.~Sturman and D.~Zeltzer.
\newblock A survey of glove-based input.
\newblock {\em {IEEE} Computer Graphics and Applications}, 14(1):30--39, 1994.

\bibitem{tang_latent_2014}
D.~Tang, H.~J. Chang, A.~Tejani, and T.-K. Kim.
\newblock Latent regression forest: Structured estimation of 3d articulated
  hand posture.
\newblock In {\em Proc. of CVPR 2014}, pages 3786--3793.

\bibitem{tang_real-time_2013}
D.~Tang, T.-H. Yu, and T.-K. Kim.
\newblock Real-time articulated hand pose estimation using semi-supervised
  transductive regression forests.
\newblock In {\em Proc. of ICCV 2013}, pages 3224--3231.

\bibitem{Tompson_TOG_2014}
J.~Tompson, M.~Stein, Y.~Lecun, and K.~Perlin.
\newblock Real-time continuous pose recovery of human hands using convolutional
  networks.
\newblock {\em ACM TOG}, 33(5):169:1--169:10, Sept. 2014.

\bibitem{Gall_GCPR_2014}
D.~Tzionas, A.~Srikantha, P.~Aponte, and J.~Gall.
\newblock Capturing hand motion with an {RGB-D} sensor, fusing a generative
  model with salient points.
\newblock In {\em Proc. of GCPR 2014}, pages 1--13.

\bibitem{wang_6d_2011}
R.~Wang, S.~Paris, and J.~Popovi\'{c}.
\newblock {6D} hands: markerless hand-tracking for computer aided design.
\newblock In {\em Proc. of UIST 2011}, pages 549--558.

\bibitem{wang_real-time_2009}
R.~Y. Wang and J.~Popovi\'{c}.
\newblock Real-time hand-tracking with a color glove.
\newblock {\em {ACM} TOG}, 28(3):63:1--63:8, July 2009.

\bibitem{wang_video-based_2013}
Y.~Wang, J.~Min, J.~Zhang, Y.~Liu, F.~Xu, Q.~Dai, and J.~Chai.
\newblock Video-based hand manipulation capture through composite motion
  control.
\newblock {\em {ACM} TOG}, 32(4):43:1--43:14, July 2013.

\bibitem{wu_capturing_2001}
Y.~Wu, J.~Lin, and T.~Huang.
\newblock Capturing natural hand articulation.
\newblock In {\em Proc. of ICCV 2001}, volume~2, pages 426--432.

\bibitem{xu_efficient_2013}
C.~Xu and L.~Cheng.
\newblock Efficient hand pose estimation from a single depth image.
\newblock In {\em Proc. of ICCV 2013}, pages 3456--3462.

\bibitem{zimmerman_hand_1986}
T.~G. Zimmerman, J.~Lanier, C.~Blanchard, S.~Bryson, and Y.~Harvill.
\newblock A hand gesture interface device.
\newblock {\em {SIGCHI} Bull.}, 17({SI}):189--192, May 1986.

\bibitem{Zollhofer:2014}
M.~Zollh\"ofer, M.~Niessner, S.~Izadi, C.~Rehmann, C.~Zach, M.~Fisher, C.~Wu,
  A.~Fitzgibbon, C.~Loop, C.~Theobalt, and M.~Stamminger.
\newblock Real-time non-rigid reconstruction using an {RGB-D} camera.
\newblock {\em ACM TOG}, 33(4):156:1--156:12, July 2014.

\end{thebibliography}
}

\end{document}